\documentclass[11pt,a4paper]{article}
\usepackage[hyperref]{emnlp2020}
\usepackage{times}
\usepackage{latexsym}

\usepackage{url}
\usepackage{graphicx}
\usepackage{amsmath}
\usepackage{booktabs}
\usepackage{algorithm}
\usepackage{algorithmic}
\usepackage{mathrsfs}
\usepackage{subfigure}
\usepackage{enumerate}
\usepackage{amsfonts}
\usepackage{multirow}
\usepackage{arydshln}
\usepackage{colortbl}
\usepackage{array}
\usepackage{setspace}
\usepackage{amssymb}
\usepackage{booktabs}
\usepackage{CJK}
\usepackage{xcolor}

\usepackage{microtype}

\aclfinalcopy 

\title{Uncertainty-Aware Semantic Augmentation for \\Neural Machine Translation}

\author{Xiangpeng Wei\textsuperscript{\rm 1,2}\Thanks{ Work done at Alibaba Group.},
	Heng Yu\textsuperscript{\rm 3},
	Yue Hu\textsuperscript{\rm 1,2}\Thanks{ Corresponding Author.}, 
	Rongxiang Weng\textsuperscript{\rm 3},
	Luxi Xing\textsuperscript{\rm 1,2},
	Weihua Luo\textsuperscript{\rm 3}\\
	\textsuperscript{\rm 1}Institute of Information Engineering, Chinese Academy of Sciences, Beijing, China\\
	\textsuperscript{\rm 2}School of Cyber Security, University of Chinese Academy of Sciences, Beijing, China\\
	\textsuperscript{\rm 3}Machine Intelligence Technology Lab, Alibaba Group, Hangzhou, China\\
	\texttt{\{weixiangpeng,huyue,xingluxi\}@iie.ac.cn} \\ \texttt{\{yuheng.yh,wengrx,weihua.luowh\}@alibaba-inc.com}\\
}

\date{}

\begin{document}
\maketitle
\begin{abstract}
As a sequence-to-sequence generation task, neural machine translation (NMT) naturally contains \textit{intrinsic uncertainty}, where a single sentence in one language has multiple valid counterparts in the other. However, the dominant methods for NMT only observe one of them from the parallel corpora for the model training but have to deal with adequate variations under the same meaning at inference. This leads to a discrepancy of the data distribution between the training and the inference phases. To address this problem, we propose \emph{uncertainty-aware semantic augmentation}, which explicitly captures the universal semantic information among multiple semantically-equivalent source sentences and enhances the hidden representations with this information for better translations. Extensive experiments on various translation tasks reveal that our approach significantly outperforms the strong baselines and the existing methods.
\end{abstract}

\section{Introduction}
In recent years neural machine translation (NMT) has demonstrated state-of-the-art performance on many language pairs with advanced architectures and large scale data~\cite{Bahdanau2014Neural,Wu2016Google,Vaswani2017Attention}. At training time the parallel data only contains one source sentence as the input and the rest reasonable ones are ignored, while at inference the resulting model has to deal with adequate variations under the same meaning. This discrepancy of the data distribution poses a formidable learning challenge of the inherent uncertainty in machine translation. Since typically there are several semantically-equivalent source sentences that can be translated to the same target sentence, but the model only observes one at training time. Thus it is natural to enable an NMT model trained with the token-level cross-entropy (CE) to capture such a rich distribution, which exactly motivates our work.

Intuitively, the NMT model should be trained under the guidance of the same latent semantics that it will access at inference time. In their seminal work, the variational models~\cite{Blunsom2008a,zhang-etal-2016-variational-neural,shah2018generative} introduce a continuous latent variable to serve as a global semantic signal to guide the generation of target translations. \citet{wei-etal-2019-translating} consider an universal topic representation for each sentence pair as global semantics for enhancing representations learnt by NMT models. \citet{yang-etal-2019-sentence} minimize the difference between the representation of source and target sentences. Although their methods yield notable results, they are still limited to one-to-one parallel sentence pairs.

To address these problems, we present a novel \emph{uncertainty-aware semantic augmentation} method, which takes account of the intrinsic uncertainty sourced from the one-to-many nature of machine translation~\cite{Ott2018Analyzing}. Specifically, we first synthesize multiple reasonable source sentences to play the role of inherent uncertainty\footnote{In our scenario, we mainly study the uncertainty in source-side, as it is problematic if the synthetic targets that inevitably contain noise and errors are used to supervise the training of models. We leave further study on this to future work.} for each target sentence. 
To achieve this, we introduce a \textit{controllable sampling} strategy to cover adequate variations for inputs, by quantifying the sharpness of the word distribution in each decoding step and taking the proper word (the one with the maximum probability if sharp enough or determined by multinomial sampling) as the output.  Then a \textit{semantic constrained network} (SCN) is developed to summarize multiple source sentences that share the same meaning into a \textit{closed semantic region}, augmented by which the model generates translations finally. By integrating such \textit{soft} correspondences into the translation process, the model can intuitively work well when fed with an unfamiliar literal expression that can be supported by its underlying semantics. In addition, given the effectiveness of leveraging monolingual data in improving translation quality~\cite{sennrich-etal-2016-improving}, we further propose to combine the strength of both 
semantic augmentation and massive monolingual data distributed in the target language.

We conduct extensive experiments in both a supervised setup with bilingual data only, and a semi-supervised setup where both bilingual and target monolingual data are available. We evaluate the proposed approach on the widely used WMT14 English$\rightarrow$French, WMT16 English$\rightarrow$German, NIST Chinese$\rightarrow$English and WMT18 Chinese$\rightarrow$English benchmarks. Experimental results show that the proposed approach consistently improves translation performance on multiple language pairs. As another bonus, by adding monolingual data in German, our approach yields an additional gain of +1.5$\sim$+3.3 BLEU points on WMT16 English$\rightarrow$German task. Extensive analyses reveal that:
\begin{itemize}
	\item Our approach demonstrates strong capability on learning semantic representations.
	\item The proposed controllable sampling strategy introduces reasonable uncertainties into the training data and generates sentences are of both high diverse and high quality.
	\item Our approach motivates the models to be consistent when processing equivalent source inputs with various lteral expressions.
\end{itemize}

\section{Preliminary}

\begin{figure*}
	\centering
	\includegraphics[width=0.85\textwidth]{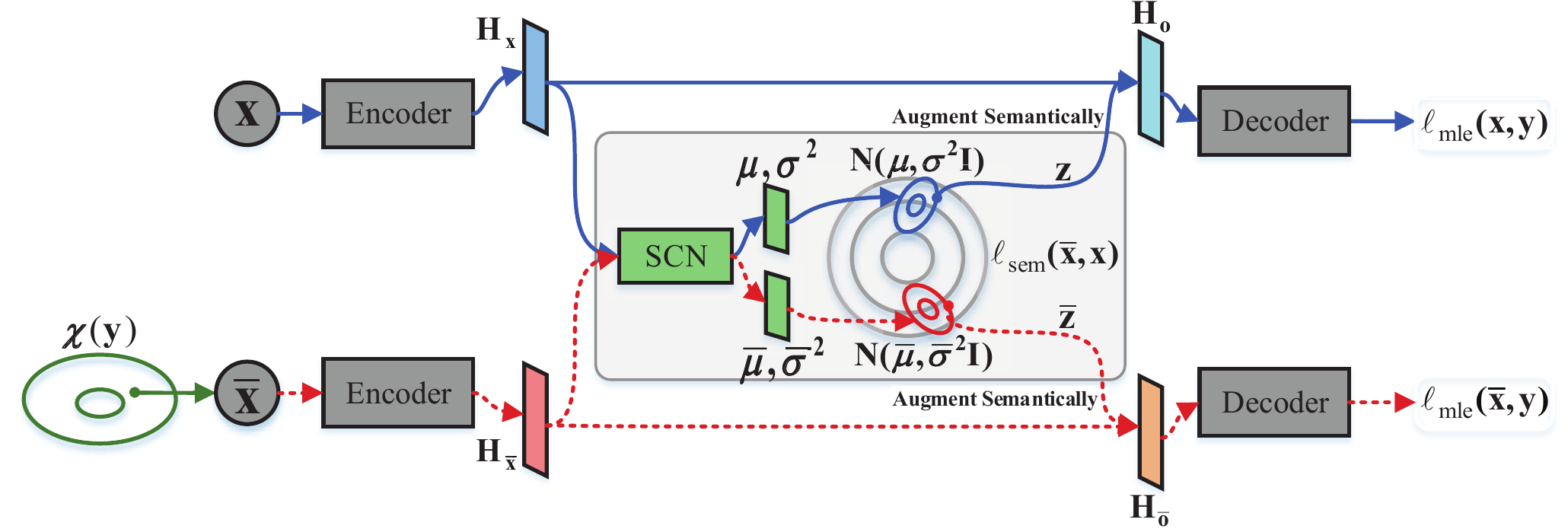}
	\caption{\label{fig:model}Uncertainty-Aware Semantic Augmentation for NMT. $\mathcal{X}({\rm \bf y})$ indicates a set of semantically-equivalent source sentences for ${\rm \bf y}$. The blue-solid and red-dashed lines represent the forward-pass information flow for $\rm \bf x$ and $\mathbf{\bar{\rm \bf x}}$, respectively. Note that our method involves a \textit{shared encoder} as well as a \textit{shared decoder} for both $\rm \bf x$ and $\mathbf{\bar{\rm \bf x}}$.
	}
\end{figure*}

Neural Machine Translation~\cite{Bahdanau2014Neural} directly models the translation probability of a target sentence ${\rm \bf y}=y_1,...,y_{T_{\rm y}}$ given its corresponding source sentence ${\rm \bf x}=x_1,...,x_{T_{\rm x}}$:
\begin{equation}
P({\rm \bf y} | {\rm \bf x}; \boldsymbol{\theta})=\prod_{i=1}^{T_{\rm y}} P(y_i | {\rm \bf y}_{<i}, {\rm \bf x};\boldsymbol{\theta}),
\label{eq:loss-intro}
\end{equation}
where $\boldsymbol{\theta}$ is a set of model parameters and ${\rm \bf y}_{<i}$ is a partial translation. The word-level translation probability is formulated as: $P(y_i | {\rm \bf y}_{<i}, {\rm \bf x};\boldsymbol{\theta}) \propto {\rm exp}\{g(y_{i-1}, s_i, c_i; {\boldsymbol{\theta}})\}$, in which $g(\cdot)$ denotes a non-linear function to predict the $i$-th target word $y_i$ from the decoder state $s_i$ and the context vector $c_i$ summarized from a sequence of representations of the encoder with an attention module. For training, given a parallel corpus $\{({\rm \bf x}^n, {\rm \bf y}^n)\}_{n=1}^{\mathcal{N}}$, the objective is to maximize ${\rm log} P({\rm \bf y}^n | {\rm \bf x}^n; \boldsymbol{\theta})$ over the entire training set.



\paragraph{Related Work on Data augmentation.} DA has been used to improve the diversity of training signals for NMT models, like randomly shuffle (swap) or drop some words in a sentence~\cite{iyyer-etal-2015-deep,artetxe2018iclr,lample2017unsupervised}, randomly replace one word in the original sentences with another word~\cite{fadaee-etal-2017-data,noising2017,kobayashi-2018-contextual,wang-etal-2018-switchout,cheng-etal-2018-towards,gao-etal-2019-soft}, syntax-aware methods~\cite{duanSyntaxDA}, as well as using target monolingual data~\cite{sennrich-etal-2016-improving,cheng-etal-2016-semi,he2016dual,Zhang2018Joint,wu-etal-2018-study,hoang-etal-2018-iterative,niu-etal-2018-bi,edunov-etal-2018-understanding,imamura-etal-2018-enhancement,xia-etal-2019-generalized}. More recently, \citet{fadaee-monz-2018-back} introduce several variations of sampling strategies targeting difficult-to-predict words. \citet{li-etal-2019-understandingDA} have studied that what benefits from data augmentation across different methods and tasks. \citet{cheng-etal-2019-robust} propose to improve the robustness of NMT models towards perturbations and minor errors by introducing adversarial inputs into training process. In contrast, we aim at bridging the discrepancy of the data distribution between the training and the inference phases, through augmenting each training instance with multiple semantically-equivalent source inputs.

\paragraph{Related Work on Uncertainty in NMT.} Recently, there are increasing number of studies investigating the effects of quantifying uncertainties in different applications~\cite{Kendall2017Bayesian,Kendall2017what,Xiao2018Quantifying,zhang-etal-2019-mitigating,zhang-etal-2019-adansp,shen-etal-2019-modelling}. However, most work in NMT has focused on improving accuracy without much consideration for the intrinsic uncertainty of the translation task itself. In their seminal work, the latent variable models~\cite{Blunsom2008a,zhang-etal-2016-variational-neural} introduce a (set of) continuous latent variable(s) to model underlying semantics of source sentences and to guide the generation of target translations. \citet{zaremoodi-haffari-2018-incorporating} propose a forest-to-sequence NMT model to make use of exponentially many parse trees of the source sentence. \citet{Ott2018Analyzing} have focused on analyzing the uncertainty in NMT that demonstrate how uncertainty is captured by the model distribution and how it affects search strategies. \cite{wang2019improving-bt} propose to quantify the confidence of NMT model predictions based on model uncertainty. Our work significantly differs from theirs. We model the inherent uncertainty by representing multiple source sentences into a closed semantic region, and use this semantic information to enhance NMT models where diverse literal expressions intuitively be supported by their underlying semantics.


\section{Uncertainty-Aware Semantic Augmentation for NMT}
 
Here, we present the \emph{uncertainty-aware semantic augmentation} (as shown in  Figure~\ref{fig:model}), which takes account of the intrinsic uncertainty of machine translation and enhances the latent representation semantically. For each sentence-pair $({\rm \bf x},{\rm \bf y})$, supposing $\mathcal{X}({\rm \bf y})$ is a set of correct source sentences for ${\rm \bf y}$, in which each sentence $\mathbf{\bar{\rm \bf x}}$ is assumed to have the same meaning as ${\rm \bf x}$. Given a training corpus $\mathcal{D}$, we introduce the objective function as:
\begin{equation}
\begin{split}
&\mathcal{J}(\boldsymbol{\theta})\\
=&\sum_{({\rm \bf x}, {\rm \bf y}) \in \mathcal{D}} \Big \{ \lambda_{1} \mathbb{E}_{\mathcal{P}_{\phi}({\rm \bf z}|{\rm \bf x})}[\underbrace{\log P({\rm \bf y} | {\rm \bf z},{\rm \bf x}; \boldsymbol{\theta})}_{\ell_{\rm mle}({\rm \bf x}, {\rm \bf y};{\rm \bf z})}]\\
&- \gamma \mathbb{E}_{\mathbf{\bar{\rm \bf x}} \sim \mathcal{X}({\rm \bf y})} \big [ \underbrace{\mathbf{KL}\big(\mathcal{P}_{\phi}(\mathbf{\bar{\rm \bf z}}|\mathbf{\bar{\rm \bf x}}) || \mathcal{P}_{\phi}({\rm \bf z}|{\rm \bf x}) \big)}_{\ell_{\rm sem}(\mathbf{\bar{\rm \bf x}}, {\rm \bf x})} \big ]\\
&+ \lambda_{2} \mathbb{E}_{\mathbf{\bar{\rm \bf x}} \sim \mathcal{X}({\rm \bf y})} \big [ \mathbb{E}_{\mathcal{P}_{\phi}(\mathbf{\bar{\rm \bf z}}|\mathbf{\bar{\rm \bf x}})}[ \underbrace{\log P({\rm \bf y} | \mathbf{\bar{\rm \bf z}}, \mathbf{\bar{\rm \bf x}}; \boldsymbol{\theta})}_{\ell_{\rm mle}(\mathbf{\bar{\rm \bf x}}, {\rm \bf y};\mathbf{\bar{\rm \bf z}})}] \big ] \Big \},
\end{split}
\label{eq:loss-x-y}
\end{equation}
where
\begin{itemize}
    \item $\ell_{\rm sem}(\mathbf{\bar{\rm \bf x}}, {\rm \bf x})$ to encourage the SCN to extract the core semantics ($\mathbf{\bar{\rm \bf z}}$ and ${\rm \bf z}$) for $\mathbf{\bar{\rm \bf x}}$ and ${\rm \bf x}$ respectively, while constraining them into a closed semantic region. It is formulated as the negative Kullback-Leibler (${\rm \bf KL}$) divergence between the semantic distributions $\mathcal{P}_{\phi}(\mathbf{\bar{\rm \bf z}}|\mathbf{\bar{\rm \bf x}})$ and $\mathcal{P}_{\phi}({\rm \bf z}|{\rm \bf x})$, where $\phi$ denotes the combined parameters of the encoder and the SCN.
    \item $\ell_{\rm mle}({\rm \bf x}, {\rm \bf y};{\rm \bf z})$ and $\ell_{\rm mle}(\mathbf{\bar{\rm \bf x}}, {\rm \bf y};\mathbf{\bar{\rm \bf z}})$ to guide the decoder to generate the output ${\rm \bf y}$ with the assist of input-invariant semantics given diverse inputs ${\rm \bf x}$ and $\mathbf{\bar{\rm \bf x}}$.
    \item $\lambda_{1}$ and $\lambda_{2}$ control the balance between the original source sentence $\rm \bf x$ and its reasonable counterparts $\mathcal{X}({\rm \bf y})$. In experiments, we set $\lambda_{1} + \lambda_{2} = 1.0$, which means a target sentence occurs once in total. $\gamma$ controls the impact of the semantic agreement training to be described in Section~\ref{sec:training}.
\end{itemize}
Intuitively, our new objective is exactly a regularized version of the widely used maximum likelihood estimation (MLE) in conventional NMT. The models are trained to optimize both the translation loss and the semantic agreement between $\rm \bf x$ and $\mathbf{\bar{\rm \bf x}}$.

In the following sections, we will first describe how to summarize multiple source sentences into a closed semantic region by developing a \textit{semantic constrained network} (SCN) in Section~\ref{sec:scn}. And then introduce the proposed \textit{controllable sampling} strategy in Section~\ref{sec:sampling} to construct adequate and reasonable variations for source inputs.

\subsection{Semantic Constrained Network}
\label{sec:scn}
\paragraph{Network Architecture.} One core component of our approach is the proposed SCN, which aims to learn the global semantics and make them no difference between multiple source sentences (${\rm \bf x}$ and $\mathbf{\bar{\rm \bf x}}$). We adopt the CNN to address the variable-length problem of a sequence of hidden representations ${\rm \bf H}_{\rm \bf x}$ (which is the output of the top encoder layer given ${\rm \bf x}$) of the encoder stack. Formally, given an encoded representation ${\rm \bf H}_{\rm \bf x}={\rm \bf H}_{{x}_1},{\rm \bf H}_{{x}_2},...,{\rm \bf H}_{{x}_{T_{\rm x}}}$, the SCN first represents it as:
\begin{equation}
\xi_{1:{T_{\rm x}}}={\rm \bf H}_{{x}_1} \oplus {\rm \bf H}_{{x}_2} \oplus ... \oplus {\rm \bf H}_{{x}_{T_{\rm x}}},
\end{equation}
where $\oplus$ is the concatenation operator to build the matrix $\xi_{1:{T_{\rm x}}}$. Then a convolution operation involves a kernel ${\rm \bf W}_{c}$ is applied to a window of $l$ words to produce a new feature:
\begin{equation}
c_i = {\rm Relu}({\rm \bf W}_{c} \otimes \xi_{i:i+l-1} + {\rm \bf b}),
\end{equation}
where $\otimes$ operator is the summation of element-wise production, ${\rm \bf b}$ is a bias term. Finally we apply a max-over-time pooling operation over the feature map $\mathbf{c}=\max\{c_1, c_2,..,c_{{T_{\rm x}}-l+1}\}$ to capture the most important feature, that is, one with the highest value. We can use various numbers of kernels with different window sizes to repeat the above process, and extract different features to form the semantic representation, denoted as ${\rm \bf H}_{\mathbf{c}}$ for $\rm \bf x$ (and ${\rm \bf H}_{\mathbf{\bar{c}}}$ for $\mathbf{\bar{\rm \bf x}}$ in a symmetric way).

\paragraph{Semantic Agreement Training.} Given the semantic distributions $\mathcal{P}_{\phi}({\rm \bf z}|{\rm \bf x})$ of $\rm \bf x$ and $\mathcal{P}_{\phi}(\mathbf{\bar{\rm \bf z}}|\mathbf{\bar{\rm \bf x}})$ of $\mathbf{\bar{\rm \bf x}}$, we formulate $\ell_{\rm sem}(\mathbf{\bar{\rm \bf x}}, {\rm \bf x})$ as the negative $\mathbf{KL}$ divergence between them:
\begin{equation}
\ell_{\rm sem}(\mathbf{\bar{\rm \bf x}}, {\rm \bf x}) = -\mathbf{KL}\big(\mathcal{P}_{\phi}(\mathbf{\bar{\rm \bf z}}|\mathbf{\bar{\rm \bf x}}) || \mathcal{P}_{\phi}({\rm \bf z}|{\rm \bf x}) \big).
\label{eq:loss-scn}
\end{equation}
We assume $\mathcal{P}_{\phi}({\rm \bf z}|{\rm \bf x})$ and $\mathcal{P}_{\phi}(\mathbf{\bar{\rm \bf z}}|\mathbf{\bar{\rm \bf x}})$ have the following forms:
\begin{equation}
\begin{split}
\mathcal{P}_{\phi}({\rm \bf z}|{\rm \bf x}) \sim \mathcal{N}(\mu,\sigma^{2} \mathbf{I}), \; \mathcal{P}_{\phi}(\mathbf{\bar{\rm \bf z}}|\mathbf{\bar{\rm \bf x}}) \sim \mathcal{N}(\bar{\mu},\bar{\sigma}^{2} \mathbf{I}).
\end{split}
\end{equation}
The mean $\mu$ ($\bar{\mu}$) and s.d. $\sigma$ ($\bar{\sigma}$) are the outputs of neural networks based on the observation ${\rm \bf H}_{\mathbf{c}}$ (or ${\rm \bf H}_{\mathbf{\bar{c}}}$), as
\begin{equation}
\begin{split}
\mu &= {\rm \bf H}_{\mathbf{c}} \cdot \mathbf{W}_{\mu} + \mathbf{b}_{\mu},\\
\log \sigma^{2} &= {\rm \bf H}_{\mathbf{c}} \cdot \mathbf{W}_{\sigma} + \mathbf{b}_{\sigma},
\end{split}
\end{equation}
or
\begin{equation}
\begin{split}
\bar{\mu} &= {\rm \bf H}_{\mathbf{\bar{c}}} \cdot \mathbf{W}_{\mu} + \mathbf{b}_{\mu},\\
\log \bar{\sigma}^{2} &= {\rm \bf H}_{\mathbf{\bar{c}}} \cdot \mathbf{W}_{\sigma} + \mathbf{b}_{\sigma},
\end{split}
\end{equation}
where $\mathbf{W}_{\mu}$, $\mathbf{b}_{\mu}$, $\mathbf{W}_{\sigma}$ and $\mathbf{b}_{\sigma}$ are trainable parameters. To obtain a representation for latent semantic distributions, we employ reparameterization technique as in~\cite{kingma2014semi,zhang-etal-2016-variational-neural}. Formally,
\begin{equation}
\begin{split}
{\rm \bf z} = \mu + \sigma \odot \epsilon, \; \mathbf{\bar{z}} = \bar{\mu} + \bar{\sigma} \odot \epsilon,
\end{split}
\label{eq:scn-output}
\end{equation}
where $\epsilon \sim \mathcal{N}(0, \mathbf{I})$ plays a role of introducing noises, and $\odot$ denotes an element-wise product. There can be other proper strategies to unify semantics of diverse inputs, we just present one example. Actually, the Gaussian form adopted here has several advantages, such as analytical evaluation of the KL divergence and ease of reparametrization for efficient gradient computation.

\paragraph{Augment Semantically.} Given the encoder output ${\rm \bf H}_{\rm \bf x}$ of ${\rm \bf x}$, we augment it semantically with the captured semantics ${\rm \bf z}$ by combining them with a gate $g = {\rm sigmoid}({\rm \bf z} \cdot {\rm \bf W}_{gz} + {\rm \bf H}_{x_t} \cdot {\rm \bf W}_{gx})$,
\begin{equation}
\begin{split}
{\rm \bf H}_{{\rm \mathbf{o}}_t} &= {\rm LayerNorm}(g \cdot {\rm \bf z} + (1-g) \cdot {\rm \bf H}_{x_t}).\\
\end{split}
\end{equation} 
Identically, ${\rm \bf H}_{\mathbf{\bar{o}}}$ can be formulated given $\mathbf{\bar{z}}$ and ${\rm \bf H}_{\mathbf{\bar{\rm \bf x}}}$. Finally, the augmented source representation ${\rm \bf H}_{\rm \mathbf{o}}$ (or ${\rm \bf H}_{\mathbf{\bar{o}}}$) is fed to the decoder to generate the final translation $\rm \bf y$ conditioned on $\rm \bf x$ (or $\mathbf{\bar{\rm \bf x}}$). In this strategy, our model can intuitively work well when meeting infrequent literal expressions as that can be pivoted by their corresponding semantic regions.

\begin{table}[t!]
	\begin{center}
		\begin{tabular}{l|l}
			\toprule[1.25pt]
			Threshold & Method\\
			\hline
			$\hbar=0$ & Multinomial sampling \\
			$\hbar=+ \infty$ & Greedy search \\
			$\hbar \in (0,+ \infty)$ & Controllable sampling \\
			\bottomrule[1.25pt]
		\end{tabular}
	\end{center}
	\caption{\label{table-controlled-sampling}Multinomial sampling and greedy search, as special cases covered in controllable sampling.}
\end{table}

\subsection{Controllable Sampling}
\label{sec:sampling}
For each target sentence ${\rm \bf y}$, we need a set of reasonable source sentences $\mathcal{X}({\rm \bf y})$ to play the role of the inherent uncertainty. Unfortunately, it is extremely cost to annotate multiple source sentences manually for tens of million target sentences. To this end, we automatically construct $\mathcal{X}({\rm \bf y})$ using a well-trained \textit{\textbf{target-to-source}} model $\overleftarrow{\boldsymbol{\theta}}$ by sampling from the predicted word distributions:
\begin{equation}
\mathbf{\bar{\rm \bf x}}_t \sim P(\cdot|\mathbf{\bar{\rm \bf x}}_{<t}, {\rm \bf y};\overleftarrow{\boldsymbol{\theta}}).
\label{eq:sampling}
\end{equation}

However, it is problematic to force the generation of a certain number of source sentences indiscriminately for each target sentence using beam search or multinomial sampling. The reason is that both of them synthesize sentences are either of less diverse or of less quality. Therefore, we propose a controllable sampling strategy to generate reasonable source sentences: at each decoding step, if the word distribution is sharp then we take the word with the maximum probability, otherwise the sampling method formulated in Eq. (\ref{eq:sampling}) is applied. Formally,
\begin{equation}
\begin{split}
&\begin{cases}
\mathbf{\bar{\rm \bf x}}_t  \sim P(\cdot|\bar{\rm \bf x}_{<t}, {\rm \bf y};\overleftarrow{\boldsymbol{\theta}}),\text{if $\varepsilon \geq \hbar$}\\
\mathbf{\bar{\rm \bf x}}_t = {\rm argmax}\big(P(\cdot|\bar{\rm \bf x}_{<t}, {\rm \bf y};\overleftarrow{\boldsymbol{\theta}})\big),\text{else}
\end{cases}
\end{split}
\label{eq:controlled-sampling}
\end{equation}
where $\varepsilon$ is exactly the information entropy respect to $P(\cdot|\mathbf{\bar{\rm \bf x}}_{<t}, {\rm \bf y};\overleftarrow{\boldsymbol{\theta}})$:
\begin{equation}
\begin{split}
\varepsilon=-\sum\nolimits_{j} \big[ & P(x^{j}|\mathbf{\bar{\rm \bf x}}_{<t}, {\rm \bf y};\overleftarrow{\boldsymbol{\theta}}) \times\\
&\log P(x^{j}|\mathbf{\bar{\rm \bf x}}_{<t}, {\rm \bf y};\overleftarrow{\boldsymbol{\theta}}) \big ],
\end{split}
\end{equation}
where $P(x^{j}|\bar{\rm \bf x}_{<t},\mathbf{y};\overleftarrow{\boldsymbol{\theta}})$ denotes the conditional probability of the $j$-th word in the vocabulary appearing after the sequence $\mathbf{x}_1,\mathbf{x}_2,...,\mathbf{x}_{t-1}$. Actually, the widely used multinomial sampling and greedy search strategies can be served as special cases of the controllable sampling. $\hbar$ is a hyperparameter that indicates the sharpness threshold of the predicted word distributions and relates our method with the special cases as shown in Table~\ref{table-controlled-sampling}. 
In practice, we repeat the above process $N$ times to generate multiple source sentences to form $\mathcal{X}({\rm \bf y})$.

\subsection{Training}
\label{sec:training}
Our framework initializes the model based on the parameters trained by the standard maximum likelihood estimation (MLE) (Eq. (\ref{eq:loss-intro})). As shown in Eq. (\ref{eq:loss-x-y}), the training objective of our approach is differentiable, which can be optimized using standard mini-batch stochastic gradient ascent techniques. To avoid the KL collapse~\cite{bowman-etal-2016-generating,zhao-etal-2017-learning}, we use a simple scheduling strategy that sets $\gamma=0$ at the beginning of training and gradually increases $\gamma$ until $\gamma=1$ is reached.

\section{Experiments}
We examine our method upon advanced \textsc{Transformer}~\cite{Vaswani2017Attention} and conduct experiments on four widely used translation tasks, including WMT14 English$\rightarrow$French (En$\rightarrow$Fr), WMT16 English$\rightarrow$German (En$\rightarrow$De), NIST Chinese$\rightarrow$English (Zh$\rightarrow$En) and WMT18 Chinese$\rightarrow$English.

\subsection{Experimental Setting}

\paragraph{Dataset} For En$\rightarrow$De, we used the WMT16\footnote{\url{http://www.statmt.org/wmt16/}} corpus containing 4.5M sentence pairs with 118M English words and 111M German words. The validation set is the concatenation of newstest2012 and newstest2013, and the results are reported on newstest2014 (test14), newstest2015 (test15) as well as newstest2016 (test16). For En$\rightarrow$Fr, we used the significantly larger WMT 2014 English-French dataset consisting of 36M sentences. The validation set is the concatenation of newstest2012 and newstest2013, and the results are reported on newstest2014 (test14). For NIST Zh$\rightarrow$En, we used the LDC\footnote{LDC2002E18, LDC2003E07, LDC2003E14, the Hansards portion of LDC2004T07-08 and LDC2005T06.} corpus consisting of 1.25M sentence pairs with 27.9M Chinese words and 34.5M English words respectively. We selected the best model using the NIST 2002 as the validation set for model selection and hyperparameters tuning. The NIST 2004 (MT04), 2005 (MT05), 2006 (MT06), and 2008 (MT08) datasets are used as test sets. For WMT18 Zh$\rightarrow$En, we used a subset of WMT18 corpus containing 8M sentence pairs. We used newsdev 2017 as the validation set and reported results on newstest 2017 as well as newtest 2018.

We used the Stanford segmenter~\cite{tseng-etal-2005-conditional} for Chinese word segmentation and applied the script \texttt{tokenizer.pl} of Moses~\cite{koehn-etal-2007-moses} for English, French and German tokenization. For En$\rightarrow$De and En$\rightarrow$Fr, all data had been jointly byte pair encoded (BPE)~\cite{sennrich-etal-2016-neural} with 32k merge operations, which results in a shared source-target vocabulary. For NIST Zh$\rightarrow$En, we created shared BPE codes with 60K operations that induce two vocabularies with 47K Chinese sub-words and 30K English sub-words. For WMT18 Zh$\rightarrow$En, we used byte-pair-encoding to preprocess the source and target sentences, forming source- and target-side dictionaries with 32K types, respectively.

\begin{table*}[t!]
	\begin{center}
		\footnotesize
		\begin{tabular}{l|c|c|cccc|cc}
			\toprule[1.25pt]
			\multirow{2}{*}{Method} & \multirow{2}{*}{Param.} &
			Training Time &
			\multicolumn{4} {c|} {NIST Zh$\rightarrow$En} & \multicolumn{2} {c} {WMT18 Zh$\rightarrow$En}\\
			\cline{4-9}
			~ & ~ & (hours) & MT04 & MT05 & MT06 & MT08 & test17 & test18\\
			\hline
			$^\dagger$\citet{Vaswani2017Attention} & 84M & 9 & 47.37 & 46.81 & 46.34 & 38.23 & 24.41 & 24.59\\
			\citet{cheng-etal-2019-robust} & N/A & N/A & 49.13 & 49.04 & 47.74 & 38.61 & N/A & N/A \\
			\hline
			{\sc Transformer} & 84M & 9 & 47.14 & 47.03 & 46.26 & 38.31 & 24.69 & 24.61\\
			{\sc Transformer$_{\rm syn}$} & 84M & $^\ddagger$10 & 47.84 & 47.90 & 47.38 & 39.64 & 25.47 & 25.06\\
			\emph{Ours} & 86M & $^\S$11.5 & \bf 49.15 & \bf 49.21 & \bf 48.88 & \bf 40.94 & \bf 26.48 & \bf 26.36\\
			\bottomrule[1.25pt]
		\end{tabular}
	\end{center}
	\caption{\label{bleu-table-ldc-zhen}BLEU [\%] on Zh$\rightarrow$En tasks. $^\dagger$ denotes replicated results using tensor2tensor (T2T) toolkit. Both the training time and the number of parameters are related to the NIST Zh$\rightarrow$En task. $^\ddagger$The time spent in synthesizing pseudo data was included. $^\S$Both the time spent in generating synthetic data and training models were included.}
\end{table*}

\paragraph{Model} We adopt the \texttt{transformer\_base} setting for Zh$\rightarrow$En translations, while both \texttt{base} and \texttt{big} settings are adopted in En$\rightarrow$De and En$\rightarrow$Fr translations. For SCN, the filter windows are set to 2, 3, 4, 5 with 128 feature maps each. We set $\hbar=2.5$, $N=3$ for balancing the translation performance and the computation complexity. During training, we set $\lambda_{1}=\lambda_{2}=0.5$, roughly 4,096 source and target tokens are paired in one mini-batch. We employ the Adam optimizer with $\beta_1=0.9$, $\beta_2=0.998$, and $\epsilon=10^{-9}$. Additionally, the same warmup and decay strategy for learning rate as~\citet{Vaswani2017Attention} is also used, with 8,000 warmup steps. For evaluation, we use beam search with a beam size of 4/5 and length penalty of 0.6/1.0 for En$\rightarrow$\{De,Fr\}/Zh$\rightarrow$En tasks respectively. We measure case-sensitive/insensitive tokenized BLEU\footnote{\url{https://github.com/moses-smt/mosesdecoder/}} by \texttt{multi-bleu.pl}/\texttt{mteval-v11b.pl} for En$\rightarrow$De/NIST Zh$\rightarrow$En, while case-sensitive detokenized BLEU is reported by the official evaluation script \texttt{mteval-v13a.pl} for WMT18 Zh$\rightarrow$En. Unless noted otherwise we run each experiment on up to four Tesla M40 GPUs and accumulate the gradients for 4 updates. For En$\rightarrow$De/NIST Zh$\rightarrow$En, each model was repeatedly run 4 times and we reported the average BLEU, while each model was trained only once on the larger WMT18 Zh$\rightarrow$En dataset. For a strictly consistent comparison, we involve two strong baselines:
\begin{itemize}
\item \textbf{\textsc{Transformer}}, which is trained on the real parallel data only.
\item \textbf{\textsc{Transformer$_{\rm syn}$}}, which \emph{is trained on the same data as ours} that consists of the real parallel data and the \emph{back-translated} corpora. The latter contains $N$ semantically-equivalent source sentences for each target sentence. These synthetic corpora are generated by a well-trained reverse NMT model using the proposed \emph{controllable sampling} (see~\ref{sec:sampling}).
\end{itemize}

\subsection{Comparison and Results}
Table~\ref{bleu-table-ldc-zhen} shows the results on Zh$\rightarrow$En tasks. For NIST Zh$\rightarrow$En, we first compare our approach with the \textsc{Transformer} model on which our model is built. As we can see, our method can bring substantial improvements, which achieves notable gain of +2.36 BLEU points on average. In addition, our best model also achieves superior results across test sets to existing systems. For a more challenging task, we also report the results on WMT18 Zh$\rightarrow$En task in Table~\ref{bleu-table-ldc-zhen}. Compared with strong baseline systems, we observe that our method consistently improves translation performance on both newstest2017 and newstest2018. These results indicate that the effectiveness of our approach cannot be affected by the size of datasets.

\begin{table}[t!]
	\begin{center}
	\footnotesize
		\begin{tabular}{l|p{0.6cm}<{\centering}p{0.6cm}<{\centering}p{0.6cm}<{\centering}|c}
			\toprule[1.25pt]
		    \multirow{2}{*}{Method} & \multicolumn{3}{c|}{En$\rightarrow$De} & En$\rightarrow$Fr \\
			~ & test14 & test15 & test16 & test14\\
			\hline	\multicolumn{5}{l}{\textit{Transformer base model}}\\
			\hline
			\citet{Vaswani2017Attention}& 27.30 & N/A & N/A & 38.10\\
			$^\ddagger$\citet{cheng-etal-2018-towards} & 28.09 & 32.47 & 36.75 & 40.21\\	\citet{cheng-etal-2019-robust} & 28.34 & N/A & N/A & N/A\\
			\hline
			\textsc{Transformer} & 27.67 & 32.04 & 36.18 & 39.86\\
			\textsc{Transformer$_{\rm syn}$} & 27.81 & 32.33 & 36.62 & 40.07\\
			\emph{Ours} & \bf 28.57 & \bf 32.95 & \bf 37.11 & \bf 41.27\\
			\hline	\multicolumn{5}{l}{\textit{Transformer big model}}\\
			\hline
			\citet{gao-etal-2019-soft} & 29.70 & N/A & N/A & N/A\\
			\citet{cheng-etal-2019-robust} & 30.01 & N/A & N/A & N/A\\
			\emph{Ours} & \bf 30.29 & \bf 34.21 & \bf 38.28 & \bf 42.92\\
			\bottomrule[1.25pt]
		\end{tabular}
	\end{center}
	\caption{\label{bleu-table-wmt-ende}BLEU [\%] on En$\rightarrow$De and En$\rightarrow$Fr translation tasks. $^\ddagger$denotes our replicated results.}
\end{table}

Table~\ref{bleu-table-wmt-ende} shows the results on WMT16 En$\rightarrow$De and WMT14 En$\rightarrow$Fr translations. For En$\rightarrow$De, when investigating semantic augmentation into NMT models, significant improvements over two baselines (up to +0.91 and +0.62 BLEU points on average respectively) can be observed. We also take existing NMT systems as comparison which use almost the same English-German corpus. Our best system outperforms the standard Transformer~\cite{Vaswani2017Attention} with +1.27 BLEU on newstest2014. It worth mentioning that our method outperforms the advanced robust NMT systems~\cite{cheng-etal-2018-towards,cheng-etal-2019-robust}, which aim to construct anti-noise NMT models, with at least +0.23 BLEU and up to +0.48 BLEU improvements. On En$\rightarrow$Fr, our method outperforms both the previous models and the in-house baselines. To further verify our approach, we study it with respect to \texttt{big} models and compare it with two related methods \cite{cheng-etal-2019-robust,gao-etal-2019-soft}. We can observe that the proposed approach achieves the best results among all methods for the same number of hidden units.

\subsection{Analysis}
\paragraph{Effect of $N$.}
To determine the number of synthetic source sentences $N$ in our system beforehand, we conduct experiments on Zh$\rightarrow$En and En$\rightarrow$De translation tasks to test how it affects the translation performance. We vary the value of $N$ from 1 to 9 with 2 as step size and the results are reported on validation sets (Table~\ref{table-N}). We can find that the translation performance achieves substantial improvement with $N$ increasing from 1 to 3. However, with $N$ set larger than 3, we get little improvement. To make a trade-off between the translation performance and the computation complexity, we set $N$ as 3 in our experiments.

\begin{table}[t!]
	\begin{center}
	\footnotesize
		\begin{tabular}{l|p{1.5cm}<{\centering}p{1.5cm}<{\centering}p{1.5cm}<{\centering}}
		\toprule[1.25pt]
			$N$ & NIST Zh$\rightarrow$En & WMT18 Zh$\rightarrow$En & WMT16 En$\rightarrow$De\\
			\hline
			1 & 43.09 & 23.01 & 24.50\\
			3 & 44.01 & 23.80 & 25.22\\
			5 & 43.92 & 23.89 & 25.30\\
			7 & 44.09 & 23.77 & 25.31\\
			9 & 44.14 & 23.68 & 25.37\\
			\bottomrule[1.25pt]
		\end{tabular}
	\end{center}
	\caption{\label{table-N}Effect of various numbers of synthetic source sentences on validation sets.}
\end{table}

\begin{table}[t!]
	\begin{center}
	\footnotesize
		\begin{tabular}{l|p{1.0cm}<{\centering}|p{1.7cm}<{\centering}p{1.7cm}<{\centering}}
		    \toprule[1.25pt]
			\multirow{4}{*}{$\hbar$} & \multirow{4}{*}{BLEU} & \multicolumn{2} {c} {Edit Distance}\\
			\cline{3-4}
			~ & ~ & \textsc{Syn} & \textsc{Syn}\\
			~ & ~ & vs. & vs. \\
			~ & ~ & \textsc{Real} & \textsc{Syn} \\
			\hline
		    \multicolumn{4} {c} {NIST Zh$\rightarrow$En} \\
		    \hline
		    BS-3 & 20.87 & 8.70 & 5.14 \\
		    \hline
			0.0 & 10.71 & 17.26 & 19.18 \\
			1.0 & 11.99 & 17.17 & 18.91 \\
			2.5 & \underline{17.60} & \underline{12.80} & \underline{12.38} \\
			4.5 & 19.47 & 9.93 & 6.24 \\
			7.0 & 20.30 & 9.07 & 4.35 \\
			\hline
		    \multicolumn{4} {c} {WMT18 Zh$\rightarrow$En} \\
			\hline
			BS-3 & 34.47 & 10.74 & 4.73\\
		    \hline
			0.0 & 24.01 & 22.55 & 21.41 \\
			1.0 & 25.22 & 22.03 & 21.09\\
			2.5 & \underline{31.29} & \underline{12.58} & \underline{12.24} \\
			4.5 & 32.96 & 9.31 & 6.29 \\
			7.0 & 33.81 & 9.37 & 5.24 \\
			\hline
		    \multicolumn{4} {c} {WMT16 En$\rightarrow$De} \\
			\hline
			BS-3 & 30.11 & 9.59 & 4.36 \\
		    \hline
			0.0 & 19.84 & 15.60 & 15.75 \\
			1.0 & 20.57 & 15.22 & 15.26 \\
			2.5 & \underline{26.44} & \underline{10.45} & \underline{10.23} \\
			4.5 & 28.07 & 8.38 & 3.95 \\
			7.0 & 29.25 & 7.29 & 2.71 \\
			\bottomrule[1.25pt]
		\end{tabular}
	\end{center}
	\caption{\label{table-h-bias}Effect of $\hbar$ on validation sets with respect to BLEU scores as well as edit distances among synthetic and real source sentences. ``\emph{BS-3}'' indicates that synthetic sentences are generated by beam search with a beam size of 3.}
\end{table}

\paragraph{Effect of $\hbar$.}
The introduction of the hyperparameter $\hbar$ aims at acquiring the proper quantity of synthetic data. To investigate the effect of it, we quantify: (1) the diversity using the edit distance among the synthetic source sentences and (2) the quality using BLEU scores of synthetic source sentences, with respect to various values of $\hbar$. 

For each target sentence in validation sets, we generate $N=3$ synthetic source sentences using \emph{controllable sampling}. Table~\ref{table-h-bias} shows the results. The BLEU scores were computed regarding the multiple synthetic sentences as a document. As in (Imamura et al., 2018), the edit distances are computed for two cases: (1) \textsc{Syn} vs. \textsc{Real}, the average distance between a synthetic source sentence (\textsc{Syn}) and the real source sentence (\textsc{Real}). (2) \textsc{Syn} vs. \textsc{Syn}, the average distance among synthetic source sentences of a target sentence (${\rm C}_{3}^{2}=3$ combinations per target sentence). We can find that when $\hbar$ tends to 0 our controlled sampling method achieves lowest BLEU scores but highest edit distances. However, if we increase $\hbar$ gradually, it can be quickly simplified to greedy search. Among all values of $\hbar$ in Table~\ref{table-h-bias}, $\hbar=2.5$ is a proper setting as it demonstrates relatively higher BLEU scores and lower word error rates (\textsc{Syn} vs. \textsc{Real}) as well as more of diversity (\textsc{Syn} vs. \textsc{Syn}) in corpora. Therefore, we set $\hbar$ as 2.5 in all of our experiments. In addition, we can observe that the controllable sampling achieves the goal of generating sentences are of both high diverse and high quality.  

\begin{table}[t!]
	\begin{center}
	\footnotesize
		\begin{tabular}{ccc|c}
			\toprule[1.25pt]
			$\ell_{\rm mle}({\rm \bf x}, {\rm \bf y})$ & $\ell_{\rm sem}$ & $\ell_{\rm mle}(\mathbf{\bar{\rm \bf x}}, {\rm \bf y})$ & BLEU\\
			\hline
			$\checkmark$ &  &  & 22.59 \\
			$\checkmark$ &  & $\checkmark$ & 23.37 \\
			$\checkmark$ & $\checkmark$ &  & 23.68 \\
			$\checkmark$ & $\checkmark$ & $\checkmark$ & 24.10 \\
			\bottomrule[1.25pt]
		\end{tabular}
	\end{center}
	\caption{\label{table-ablation}Ablation study on WMT18 Zh$\rightarrow$En validation set. ``$\checkmark$'' means the loss function is included in the training objective.}
\end{table}

\paragraph{Ablation Study.}
We perform an ablation study of our training objective formulated in Eq. (\ref{eq:loss-x-y}) that contains three loss items. As shown in Table~\ref{table-ablation}, the translation performance decreases by 0.42 BLEU points when removing $\ell_{\rm mle}(\mathbf{\bar{\rm \bf x}}, {\rm \bf y})$ while that increases to 0.73 BLEU points when $\ell_{\rm sem}$ is excluded. In addition, only adding $\ell_{\rm sem}$ is able to achieve an improvement of +1.09 BLEU points.

\begin{table}[t!]
	\begin{center}
	\footnotesize
		\begin{tabular}{l|c|c}
		    \toprule[1.25pt]
			\textit{Ours} Method & $N$ & BLEU\\
			\hline
			w/ beam search & 3 & 23.41\\
			w/ Multinomial sampling & 3 & 23.74\\
			w/ Controllable sampling & 3 & 24.10\\
			\bottomrule[1.25pt]
		\end{tabular}
	\end{center}
	\caption{\label{bleu-table-cs}Effect of different methods to generate multiple synthetic data. Experiments are conducted on WMT18 Zh$\rightarrow$En validation set.}
\end{table}

\paragraph{Effect of Controllable Sampling}
The widely used multinomial sampling and beam (greedy) search can be viewed as two special cases of the newly introduced controllable sampling. As in Table~\ref{bleu-table-cs}, our controllable sampling method achieves the best result among them on the validation set. We think that reasonable uncertainties can be mined via our controllable sampling strategy.

\begin{figure}[t!]
	\centering
	\includegraphics[width=0.5\textwidth]{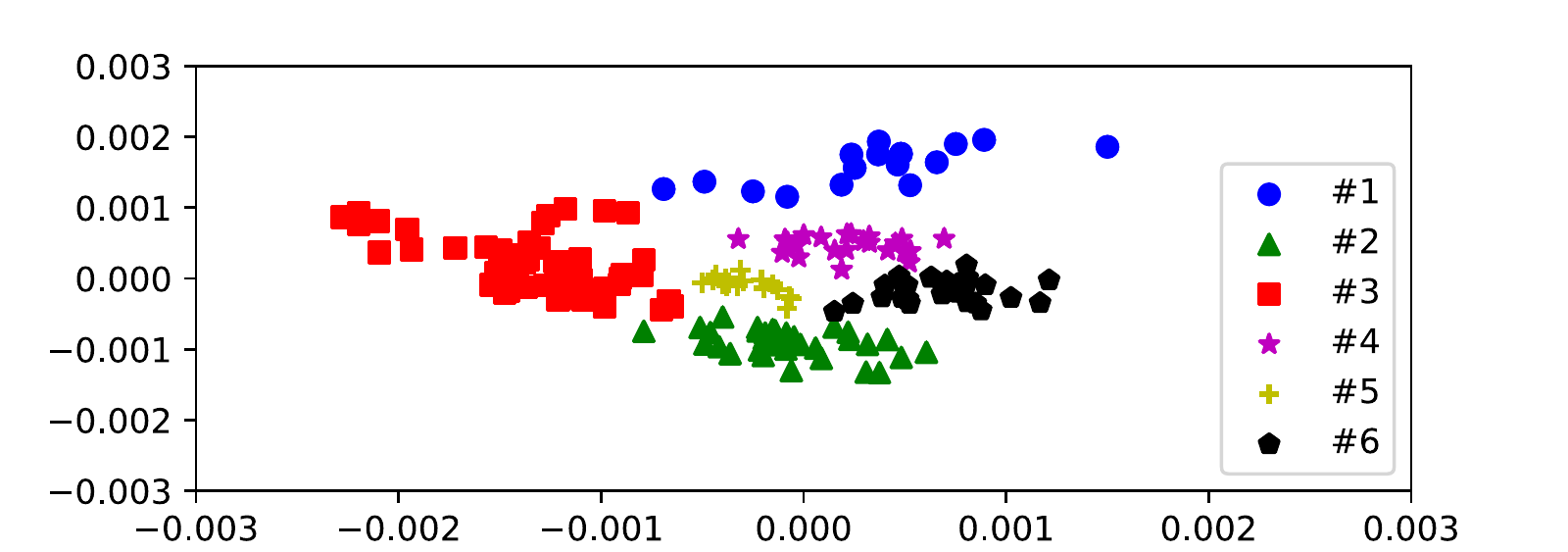}
	\caption{\label{fig:pca}The figure shows a 2-dimensional PCA projection of the semantic representations (${\rm \bf z}$ in Eq. (\ref{eq:scn-output})) for six examples in the training corpus.}
\end{figure}

\paragraph{Visualization of Latent Space.}
We would like to verify whether our approach can capture semantics. Fortunately, there are such cases in the training set: a target sentence appears several times with different source sentences. We take some of them as examples, in which there are at least 17 unique source sentences for each target sentence. We visualize the semantic representations captured by the SCN of these examples in Figure~\ref{fig:pca}. We observe that the representations are clearly clustered into 6 groups as expected, although demonstrating some noises, which 
reveal the strong capability of our approach to capture semantic representations.

\begin{table}[t!]
	\footnotesize
	\centering
	\begin{tabular}{p{1.5cm}<\centering|p{5.4cm}}
		\toprule[1.25pt]
		Input \#1 & \begin{CJK*}{UTF8}{gbsn}
			我认为我们可以重新启动这些品牌，而且现在时间正合适。
		\end{CJK*}\\
		\hline
		\textsc{Trans$_{\rm syn}$} & I think we can \textcolor{blue}{restart} these brands, and the time is right.\\
		\hline
		\emph{Ours} & I think we can \textcolor{red}{relaunch} these brands, and now is the right time.\\
		\midrule[1.25pt]
		Input \#2 & \begin{CJK*}{UTF8}{gbsn}
			我想现在是时候重新发布这些品牌了。
		\end{CJK*}\\
		\hline
		\textsc{Trans$_{\rm syn}$} & I think it is time to \textcolor{blue}{reissue} these brands.\\
		\hline
		\emph{Ours} & I think it's time to \textcolor{red}{relaunch} these brands. \\
		\midrule[1.25pt]
		Input \#3 & \begin{CJK*}{UTF8}{gbsn}
			我认为我们可以重新上新这些品牌，而且现在时间正合适。
		\end{CJK*}\\
		\hline
		\textsc{Trans$_{\rm syn}$} & I think we can \textcolor{blue}{renew} these brands, and now is the right time.\\
		\hline
		\emph{Ours} & I think we can \textcolor{red}{re-launch} these brands, and the time is right now.\\
		\bottomrule[1.25pt]
	\end{tabular}
	\caption{\label{tb:example}Translation examples of \textsc{Transformer$_{\rm syn}$} (\textsc{Trans$_{\rm syn}$} for short) and our method on various inputs under the same meaning on WMT18 Zh$\rightarrow$En.}
\end{table}

\paragraph{Case Study.}
Table~\ref{tb:example} shows an example translation (more examples are shown in the Appendix C). In this example, input \#1, \#2 and \#3 have the same meaning. For input \#1 and \#3, ``\begin{CJK*}{UTF8}{gbsn}启动\end{CJK*}'' and ``\begin{CJK*}{UTF8}{gbsn}上新\end{CJK*}'' both mean ``launch new products'' here. The input \#2 presents a different literal expression. Compared to \textsc{Transformer$_{\rm syn}$}, our approach motivates the models to be consistent when processing equivalent source inputs with various lteral expressions.

\subsection{Semi-supervised Setting}
Given the effectiveness of leveraging monolingual data in improving translation quality~\cite{sennrich-etal-2016-improving}, we further propose to improve our proposed model using target monolingual data on WMT16 En$\rightarrow$De translation. Specifically, we augment the original parallel data of WMT16 corpus containing 4.5M sentence pairs by 24M\footnote{Due to resource constraints, we use a subset with randomly selected 24M sentences of German monolingual newscrawl data distributed with WMT18, instead of the whole corpora (scale to 226M) used in~\cite{edunov-etal-2018-understanding}.} unique sentences randomly extracted from German monolingual newscrawl data. All of them are no longer than 100 words after tokenizing and BPE processing. We synthesize multiple source sentences for each monolingual sentence via controllable sampling (Section~\ref{sec:sampling}), and the one with the highest probability is served as the \emph{real} source sentence (i.e., $\rm \bf x$). We upsample the parallel data with a rate of 5 so that we observe every bitext sentence 5 times more often than each monolingual sentence. The resulted data is finally used to re-train our models and perform 300K updates on 8 P100 GPUs. Due to resource constraints, we adopt the smaller \texttt{transformer\_base} setting here.

\begin{table}[t!]
	\begin{center}
	\footnotesize
		\begin{tabular}{l|c|c|c}
			\toprule[1.25pt]
			Method& test14 & test15 & test16\\
			\hline
			\citet{wang2019improving-bt}, \texttt{big} & 31.00 & 32.01 & N/A\\
			\citet{edunov-etal-2018-understanding}, \texttt{big} & \bf 35.00 & 34.87 & 37.89\\
			\hline
			\textsc{Transformer}, \texttt{base} & 27.67 & 32.04 & 36.18 \\
			+ Monolingual Data & 30.14 & 34.17 & 37.28\\
			\hline
			\emph{Ours}, \texttt{base} & 28.57 & 32.95 & 37.11\\
			+ Monolingual Data & 31.87 & \bf 35.19 & \bf 38.65\\
			\bottomrule[1.25pt]
		\end{tabular}
	\end{center}
	\caption{\label{bleu-table-mono}BLEU scores [\%] on WMT16 En$\rightarrow$De test sets (newstest2014$\sim$2016) with monolingual data. \citet{wang2019improving-bt} used 2M extra back-translated data and \citet{edunov-etal-2018-understanding} used 226M German monolingual sentences during back-translation.}
\end{table}

Table~\ref{bleu-table-mono} summarizes our results and compares to other work in the literature. After incorporating monolingual data, our method yields an additional gain of +1.5$\sim$+3.3 BLEU points. For comparison, \citet{wang2019improving-bt} quantify the prediction confidence using model uncertainty to alleviate the noisy back-translated parallel data and achieve 31 BLEU on newstest2014. \citet{edunov-etal-2018-understanding} achieve as high as 35.0 BLEU on newstest2014 by adopting the \texttt{transformer\_big} setting and relying on massive (scale to 226M) monolingual data. For comparison, our models fall behind \citet{edunov-etal-2018-understanding}'s method on newstest2014 but achieve superior results on other two test sets. This reveals that the proposed method is surprisingly effective and complements existing non-semantic data augmentation techniques quite well.

\section{Conclusion and Future Work}
We present an uncertainty-aware semantic augmentation method to bridge the discrepancy of the data distribution between the training and the inference phases for dominant NMT models. In particular, we first synthesize a proper number of source sentences to play the role of intrinsic uncertainties via the controllable sampling for each target sentence. Then, we develop a semantic constrained network to summarize multiple source inputs into a closed semantic region which is then utilized to augment latent representations. Experiments on WMT14 English$\rightarrow$French, WMT16 English$\rightarrow$German, NIST Chinese$\rightarrow$English and WMT18 Chinese$\rightarrow$English translation tasks show that the proposed method can achieve consistent improvements across different language pairs.

While we showed that uncertainty-aware semantic augmentation with Gaussian priors is effective, more work is required to investigate if such an approach will also be successful for
more sophisticated priors. In addition, learning universal representations among semantically-equivalent source and target sentences \cite{wei-etal-2020-on} can complete the proposed method.

\section*{Acknowledgments}
We would like to thank all of the anonymous reviewers for their invaluable suggestions and helpful comments. This work is supported by the National Key Research and Development Programs under Grant No. 2017YFB0803301, No. 2016YFB0801003 and No. 2018YFB1403202.

\bibliographystyle{acl_natbib}
\bibliography{anthology,emnlp2020}

\end{document}